\documentclass[sigconf]{acmart}

\AtBeginDocument{%
 }

\copyrightyear{2025}
\acmYear{2025}
\setcopyright{acmlicensed}
\acmConference[CIKM '25] {Proceedings of the 34th ACM International Conference on Information and Knowledge Management}{ November 10--14, 2025}{Seoul, Republic of Korea.}
\acmBooktitle{Proceedings of the 34th ACM International Conference on Information and Knowledge Management (CIKM '25), November 10--14, 2025, Seoul, Republic of Korea}
\acmISBN{979-8-4007-2040-6/2025/11}
\acmDOI{10.1145/3746252.3761621}

\makeatletter
\gdef\@copyrightpermission{
  \begin{minipage}{0.3\columnwidth}
   \href{https://creativecommons.org/licenses/by/4.0/}
{\includegraphics[width=0.90\textwidth]{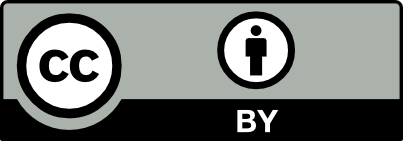}}
  \end{minipage}\hfill
  \begin{minipage}{0.7\columnwidth}
   \href{https://creativecommons.org/licenses/by/4.0/}{This work is licensed under a Creative Commons Attribution International 4.0 License.}
  \end{minipage}
  \vspace{5pt}
}
\makeatother

\author{Ming Cheng}
\affiliation{%
  \institution{Virginia Tech}
  \city{Blacksburg}
  \state{Virginia}
  \country{USA}
}
\email{ming98@vt.edu}

\author{Tong Wu}
\affiliation{%
  \institution{Virginia Tech}
  \city{Blacksburg}
  \state{Virginia}
  \country{USA}
}
\email{tongw@vt.edu}

\author{Jiazhen Hu}
\affiliation{%
  \institution{Virginia Tech}
  \city{Blacksburg}
  \state{Virginia}
  \country{USA}
}
\email{hjiazhen@vt.edu}

\author{Jiaying Gong}
\affiliation{%
  \institution{Virginia Tech}
  \city{Blacksburg}
  \state{Virginia}
  \country{USA}
}
\email{gjiaying@vt.edu}

\author{Hoda Eldardiry}
\affiliation{%
  \institution{Virginia Tech}
  \city{Blacksburg}
  \state{Virginia}
  \country{USA}
}
\email{hdardiry@vt.edu}

\usepackage[ruled,vlined]{algorithm2e}
\usepackage{amsmath}

\PassOptionsToPackage{table}{xcolor}
\usepackage{xcolor}
\usepackage{multirow}

\begin{document}

\title{VideoAVE: A Multi-Attribute Video-to-Text Attribute Value Extraction Dataset and Benchmark Models}


\begin{abstract}
Attribute Value Extraction (AVE) is important for structuring product information in e-commerce. However, existing AVE datasets are primarily limited to text-to-text or image-to-text settings, lacking support for product videos, diverse attribute coverage, and public availability. To address these gaps, we introduce VideoAVE, the first publicly available video-to-text e-commerce AVE dataset across 14 different domains and covering 172 unique attributes. To ensure data quality, we propose a post-hoc CLIP-based Mixture of Experts filtering system (CLIP-MoE) to remove the mismatched video-product pairs, resulting in a refined dataset of 224k training data and 25k evaluation data. In order to evaluate the usability of the dataset, we further establish a comprehensive benchmark by evaluating several state-of-the-art video vision language models (VLMs) under both attribute-conditioned value prediction and open attribute-value pair extraction tasks. Our results analysis reveals that video-to-text AVE remains a challenging problem, particularly in open settings, and there is still room for developing more advanced VLMs capable of leveraging effective temporal information. The dataset and benchmark code for VideoAVE are available at: ~\url{https://github.com/gjiaying/VideoAVE}.

\end{abstract}



\maketitle

\section{Introduction}
\begin{figure}[htp] 
 \center{\includegraphics[height=3.7cm,width=7.4cm]{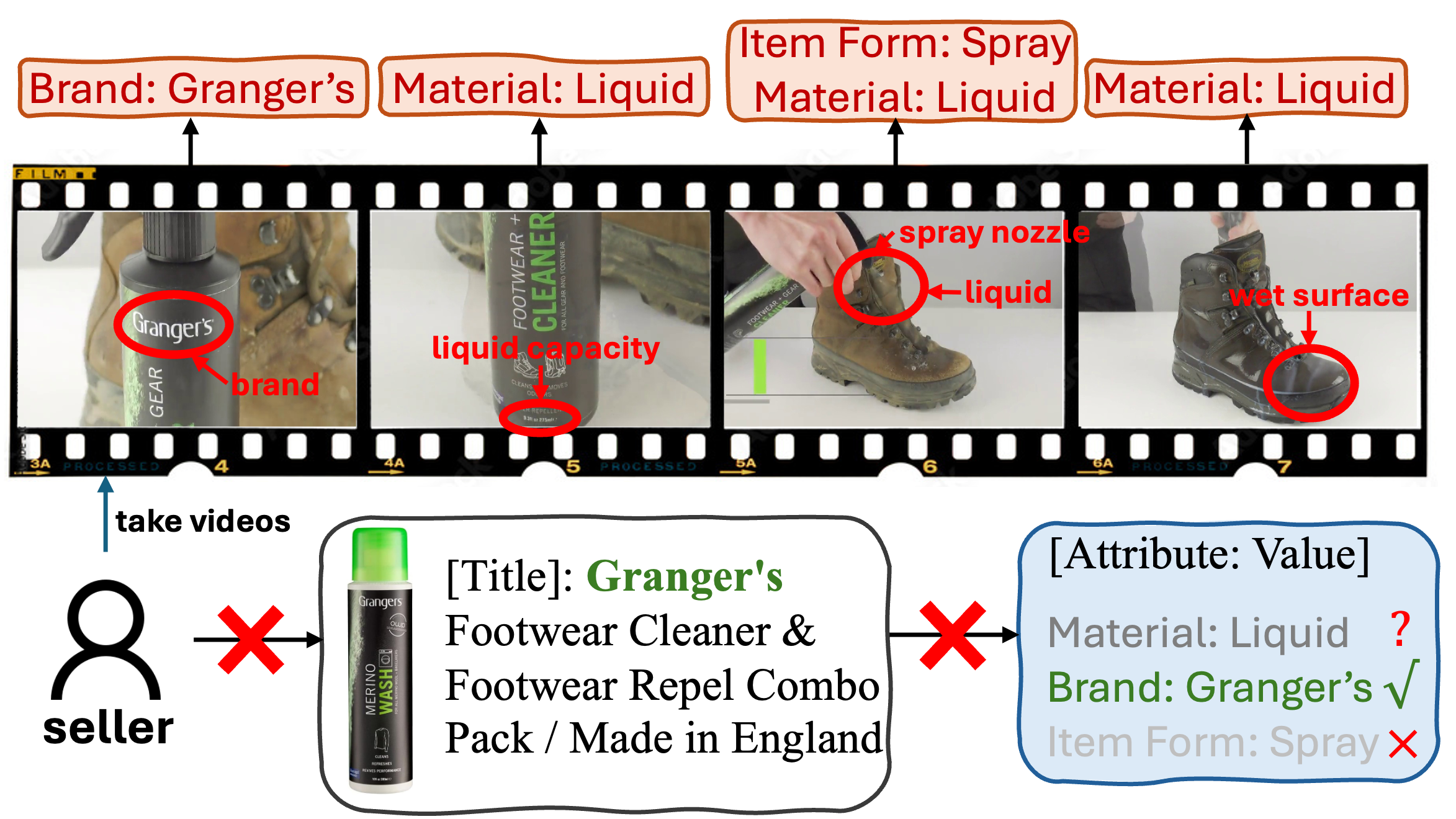}}
 \caption{\label{fig:example} Comparison of product attribute value generation from videos vs. from static images and titles.}
 \vspace{-4mm}
 \end{figure}

Attribute Value Extraction (AVE) aims at generating product attribute values from the product information.
However, existing AVE datasets and benchmarks suffer from several limitations.
First, existing datasets are limited to text-to-text or image-to-text settings. 
Early AVE datasets such as OpenTag~\cite{zheng2018opentag}, AdaTag~\cite{yan2021adatag}, OA-Mine~\cite{zhang2022oa}, e-commerce5PT~\cite{shrimal2022ner}, MAVE~\cite{yang2022mave}, WDC-PAVE~\cite{brinkmann2024using}, AE-110K, and AE-650K~\cite{xu2019scaling} primarily focus on text-only product information.
As multimodal approaches gain prominence, subsequent AVE datasets begin incorporating visual elements. 
For example, MAE~\cite{logan2017multimodal}, MEPAVE~\cite{zhu2020multimodal}, DESIRE~\cite{zhang2023pay}, and recent ImplicitAVE~\cite{zou2024implicitave} expand multimodal capabilities by combining product images with text.
Though some implicit values that are never mentioned in the product title can be inferred from images, many attribute values remain challenging to predict accurately.
For example, in Figure~\ref{fig:example}, `Item Form: Spray' can not be reliably inferred from the product image with the product title. However, this information becomes apparent in the product video, where the spray nozzle and the visible wet surface of the shoe clearly indicate a liquid material and spray form.
This highlights the advantage of video input in revealing dynamic and implicit attribute cues that are difficult to capture.
Second, existing AVE datasets~\cite{zheng2018opentag, zhu2020multimodal, zheng2018opentag, shrimal2022ner, zhang2022oa} contain only a limited set of aspects as label classes.
Thus, prior works have primarily adopted classification-based models~\cite{10.1145/3583780.3615142, 10.1145/3539597.3570423, ghosh2023d}.
While the most recent AVE studies~\cite{gong-etal-2025-mice, zou2024implicitave, fang2024llm} have shifted toward generation-based approaches using VLMs, they still limit the value space by explicitly specifying possible value choices within the prompts.

To address these issues, we present VideoAVE, the first video-to-text AVE dataset. 
A comparison of VideoAVE with existing AVE datasets are shown in Table~\ref{tab:dataset_comparison}.
We initially source the data from the Amazon Review Dataset~\cite{hou2024bridging} by only keeping the data with MP4-format videos, multiple attributes, and video-perceivable attributes (attributes can be derived from videos).
Then we curate the dataset using our proposed post-hoc CLIP-based Mixture of Experts (CLIP-MoE) filtering mechanism to address the inconsistency between product videos and their associated titles.
This yields a more refined and quality-improved multi-label video-to-text AVE dataset of 224k training and 25k evaluation data spanning 14 diverse domains with an average of 3.43 attributes for each product and a total of 172 unique attributes.
Statistics of VideoAVE are shown in Table~\ref{tab:dataset_statistics}.

As VLMs~\cite{li2023blip, liu2023visual, zhu2023minigpt, wu2024visionllm, shi2024llamafusion, peng2023kosmos, wu2024liquid} have extended their strong performance on image-text understanding to the video domain~\cite{alayrac2022flamingo, li2023videochat, maaz2023video, yang2024qwen2, jin2024video, wang2025internvideo2, kondratyuk2023videopoet, wang2024gpt4video} and there is no existing work exploring video VLMs on the AVE task, we establish VideoAVE, a new benchmark dataset, and a comprehensive evaluation framework on our proposed VideoAVE.
We conduct a systematic evaluation of four state-of-the-art open-source video MLLMs in both zero-shot setting and fine-tuned settings.
We evaluate performance across multiple settings including attribute-conditioned value prediction, open attribute-value pair extraction, category-level and attribute-level performance.
We also compare the impact of different input modalities including text, image and video to explore how temporal dynamics in video contribute to AVE performance.
Our findings show that while video input offers unique advantages by capturing dynamic visual cues, video-to-text AVE remains a challenging task for current open-source MLLMs, particularly in the open attribute–value pair extraction scenario, which closely reflects real-world conditions.

\setlength{\tabcolsep}{3pt}

\begin{table}[]
\centering
\caption{\label{tab:dataset_comparison} Comparison of existing AVE datasets.}
\scriptsize
\begin{tabular}{lccccccc}
\hline
Dataset & \begin{tabular}[c]{@{}c@{}}Video\\Modality\end{tabular} & \begin{tabular}[c]{@{}c@{}}Multi-\\label\end{tabular} & \begin{tabular}[c]{@{}c@{}}Open-\\source\end{tabular} & \begin{tabular}[c]{@{}c@{}}Rigorous\\Filtering\end{tabular} & \begin{tabular}[c]{@{}c@{}}Multi-\\Domain\end{tabular} & Language & Size \\
\hline
OpenTag \citep{zheng2018opentag} & $\times$ & \checkmark & $\times$ & $\times$ & \checkmark & English & $-$ \\
MAE \citep{logan2017multimodal} & $\times$ & \checkmark & \checkmark & $\times$ & \checkmark & English & $7.6M$ \\
AE-650K \citep{xu2019scaling} & $\times$ & $\times$ & \checkmark & $\times$ & \checkmark & Chinese & $657.4K$ \\
MEPAVE \citep{zhu2020multimodal} & $\times$ & \checkmark & \checkmark & $\times$ & \checkmark & Chinese & $87.2K$ \\
AdaTag \citep{yan2021adatag} & $\times$ & \checkmark & $\times$ & $\times$ & $\times$ & English & $333K$ \\
MAVE \citep{yang2022mave} & $\times$ & \checkmark & \checkmark & $\times$ & \checkmark & English & $2.2M$ \\
DESIRE \citep{zhang2023pay} & $\times$ & \checkmark & $\times$ & $\times$ & $\times$ & Chinese & $\sim100K$ \\
e-commerce5PT \citep{shrimal2022ner} & $\times$ & \checkmark & $\times$ & $\times$ & \checkmark & English & $295.6K$ \\
OA-Mine \citep{zhang2022oa} & $\times$ & \checkmark & \checkmark & \checkmark & \checkmark & English & $750K$ \\
WDC-PAVE \citep{brinkmann2024using} & $\times$ & \checkmark & \checkmark & \checkmark & \checkmark & English & $4.7K$ \\
ImplicitAVE \citep{zou2024implicitave} & $\times$ & $\times$ & \checkmark & $\times$ & \checkmark & English & $70.2K$ \\
\hline
VideoAVE (Ours) & \checkmark & \checkmark & \checkmark & \checkmark & \checkmark & English & $248.8K$ \\
\hline
\end{tabular}
\end{table}

\section{Dataset Construction}
\subsection{Initial Data Collection}
We source product information, including product IDs, video URLs, titles, and attribute-values, from the Amazon Review Dataset~\cite{hou2024bridging}, which is the largest publicly available multi-modal, multi-domain dataset.
However, this dataset is impractical for the task of video-to-text product attribute value generation as it has the following limitations: 
(1) It is not feasible to extract all product attributes from video content because some attributes, such as size or quantity, are particularly challenging to derive from purely visual information;
(2) The actual product or its details are inconsistent with the content shown in the video; 
(3) The attribute values in the raw dataset are not explicitly presented, with a lack of cleanliness and clarity.

\subsection{Data Curation by MoE}
\subsubsection{Task-Oriented Data Pruning} 
We first refine the sourced data by removing data that does not have corresponding video content.
Given the nature of the task of inferring product attribute values from visual content, it is important to ensure that the dataset only includes attributes that are realistically derivable from video signals. 
Thus, we further manually inspect and remove attributes that:
\begin{itemize}
    \item values include specific numbers and are almost impossible to infer from visual modality, such as `Extension Length: 14 Inches', `Noise: 56 dB', `Item Weight: 1.13 Pounds', etc.
    \item values are subjective and ambiguous, such as `Theme: Sky, Leopard, Starry', `Special Feature: Lightweight, Non Slip, Adjustable', `Style: 3 Bundle Closure', etc.
    \item values are simple Yes/No questions, such as `Batteries Required?: No', `Is Dishwasher Safe?: Yes', `Remote Control Included?: Yes', `Is Autographed?: No', etc.
    \item values are long descriptions, such as `Warranty Description', `Product Care Instruction', `Return Policy'. etc.
\end{itemize}
We finalize the attribute list by prompting with GPT-4 and human inspection, with the prompt: 
\textit{`Please identify the commonly used attributes from the \{category\} that can be visually detected in product videos based on the \{attribute\_list\}:'}.
We remove the data that only includes a single attribute-value pair to ensure a multi-label dataset. 

\subsubsection{Post-hoc CLIP-MoE Filtering}
For contrastive VLM pretraining, models are trained to align videos with captions that accurately describe their content. 
To improve the data quality and robustly address the inconsistency between product videos and their metadata, we design a post-hoc CLIP-based Mixture of Experts (CLIP-MoE) filtering mechanism. This approach filters out mismatched video-title pairs by collectively considering semantic similarity scores computed across multiple vision-language models, ensuring that only data with strong and consistent video-text alignment is retained.

Let $\mathcal{D} = \{(v_i, t_i)\}_{i=1}^N$ denote the original dataset consisting of $N$ video-title pairs, where $v_{i}$ is the product video and $t_{i}$ is the product title. 
Each data sample $i$ results in a similarity score vector:
\begin{equation}
\mathbf{s}_i = [s_i^{(1)}, s_i^{(2)}, \ldots, s_i^{(M)}] \in \mathbb{R}^M
\end{equation}
where $M$ is the number of contrastive pre-trained vision-language models, and $s_i^{(m)} \in \mathbb{R}$ is the similarity score between $v_{i}$ and $t_{i}$~\footnote{$M=3$, CLIP-MoE leverages models of X-CLIP~\cite{ni2022expanding}, ViCLIP~\cite{wang2023internvid}, and VideoCLIP~\cite{askvideos2024videoclip}}.
To mitigate distributional discrepancies, z-score normalization is applied to $s_i^{(m)}$:
\begin{equation}
\tilde{s}_i^{(m)} = \frac{s_i^{(m)} - \mu^{(m)}}{\sigma^{(m)}}, \quad \text{for } m = 1, \ldots, M
\end{equation}
where $\mu^{(m)}$ and $\sigma^{(m)}$ denote the mean and standard deviation of the m-th model's similarity scores.
The final filtered dataset $\mathcal{D}'$ is:
\begin{equation}
\mathcal{D}' = \left\{(v_i, t_i) \in \mathcal{D} \mid \sum_{m=1}^M \mathbb{I}(\tilde{s}_i^{(m)} > \tau) \geq K \right\}
\end{equation}
where $\mathbb{I}(\cdot)$ is the indicator function, $\tau$ is the threshold for filtering, and $K$ is the number of models which scores are higher than $\tau$.


\begin{table*}[]
\small
\centering
\caption{\label{tab:main_result} Category-level fuzzy F1 (\%) results under the attribute-conditioned setting and the generalized setting. Best results among pre-trained models are reported in bold. Model highlighted in grey background indicates fine-tuned performance.}
\begin{tabular}{lcccccccccccccc}
\hline
Model & Appl. & Arts & Auto  & Baby & Beauty  & Clothes & Groc. & Indus. & Music  & Patio  & Pet & Phones & Sports & Toys  \\ \hline
\multicolumn{15}{c}{Attribute-Conditioned Setting}                 \\ \hline
 Video-LLaVA     &  15.96      &   11.68       &   13.26    &    12.53        &  10.97     &   19.73         &  10.16          & 14.04      &  15.16     &  12.93       &    15.09        &   13.02     &   14.56      &   15.26    \\
 VideoLLaMA3     & 33.33  & 28.72    & 29.22 & 33.44      & 29.20 & 35.93      & 25.97      & 31.35 & 31.84 & 29.56   & 29.49      & 30.34  & 31.89   & 33.10 \\
 InternVideo2.5     &  \textbf{39.22}      &  \textbf{31.67}        &   32.22    &   35.22         &  32.01     &  34.93          &    26.87        &  \textbf{35.24}     &   37.46    &   32.18      &    30.82        &  35.17      &   33.61      &  33.94     \\
 Qwen2.5-VL     &  39.18      &   31.58       & \textbf{32.40}      &  \textbf{37.63}          &  \textbf{33.23}     &   \textbf{42.51}         &    \textbf{31.01}        & 34.27      &  \textbf{37.75}     &   \textbf{33.08}      &    \textbf{30.88}        &   \textbf{35.19}     &   \textbf{36.40}      &  \textbf{36.86}     \\ 
 Qwen2.5-VL$^{*}$     &  \textbf{62.15}      &   \textbf{52.81}       & \textbf{55.63}     &  \textbf{63.80}          &  \textbf{57.34}     &   \textbf{61.41}         &    \textbf{56.29}        & \textbf{56.98}      &  \textbf{64.08}     &   \textbf{59.89}      &    \textbf{53.20}        &   \textbf{66.15}     &   \textbf{63.98}      &  \textbf{59.39}     \\
 \hline
 \multicolumn{15}{c}{Generalized Setting}                 \\ \hline
  Video-LLaVA     &  5.65      &   5.36       &   6.47    &    6.95        &  4.69     &   9.13         &  3.27          & 6.48      &  6.09     &  6.27       &    4.70        &   5.81     &   7.21      &   5.76    \\
 VideoLLaMA3     & 9.23  & 8.70    & 8.14 & 11.04      & 8.03 & 12.35      & 4.75      & 8.96 & 8.33 & 9.18   & 7.84      & 10.06  & 10.59   & 8.30 \\
 InternVideo2.5     &  \textbf{11.95}      &  \textbf{12.32}        &   \textbf{10.16}    &   \textbf{14.76}         &  \textbf{11.37}     &  \textbf{16.40}          &    \textbf{8.22}        &  \textbf{11.89}     &   \textbf{9.93}    &   \textbf{10.73}      &    \textbf{9.07}        &  \textbf{14.51}      &   \textbf{13.29}      &  \textbf{9.73}     \\
 Qwen2.5-VL     &  4.66      &   7.24       & 5.61      &  9.27          &  6.47     &   10.79         &    6.15        & 5.64      &  8.33     &   6.21      &    5.05        &   10.06     &   7.82      &  5.65     \\ 
  Qwen2.5-VL$^{*}$     &  \textbf{34.50}      &   \textbf{32.40}       & \textbf{31.36}     &  \textbf{38.18}          &  \textbf{38.12}     &   \textbf{42.44}         &    \textbf{38.49}        & \textbf{36.43}      &  \textbf{41.86}     &   \textbf{35.01}      &    \textbf{29.48}        &   \textbf{43.40}     &   \textbf{39.81}      &  \textbf{34.29}     \\
 \hline
\end{tabular}
\end{table*}

\subsection{Dataset Statistics}
\begin{table}[]
\small
\centering
\caption{\label{tab:dataset_statistics} Dataset statistics across 14 different domains.}
\begin{tabular}{l|ccccccc}
\hline
   Domain        & \#Raw & \#MP4 & \#Filter & \#Train & \#Test & \#Ave. & \#Uni. \\ \hline
Appl. &   94.3k    &  2598     &   1925         &   1733      &   192     &     3.68          &     22           \\
Arts       &  801.4k     &   19579    &    13671        &   12304      &  1367      &     3.39          &   35             \\
Auto       &  2.0M     &  8083     &    5740        &      5166   &    574    &   3.25            &   54             \\
Baby       &   217.7k    &   13209    &   9581         &   8625      &   956     &      3.76         &      37          \\
Beauty     &    1.1M   &   43625    &    35030        &31527         &   3503     &   3.16            & 32               \\
Clothes    &  7.2M     &   17255    &  12640          &  11376       &   1264     &   3.42            &     26           \\
Groc.    &  603.3k     &  5993     &   4006         &    3606     &  400      &      2.46         &   18         \\
Indus.   & 427.6k      &   12190    &  10515          &    9463     &    1052    &    3.28           &  42              \\
Music      &   213.6k    &   26285    &    19292        &    17363     &   1929     &   3.54            &   28             \\
Patio      &   851.9k    &   40947    &   29506         &   26555      &   2951     &   3.77            &  64              \\
Pet        &   111.5k    &   27862    &    23734        &  21358       &   2367     &    3.14           &    28            \\
Phones     &      710.5k  &  38483     &   27113         &     24402    &   2711     &      3.22         &  26              \\
Sports     &    1.6M   &   50558    &   36471         &            32824    &    3647       &  4.05   &   71             \\
Toys       & 890.9k      &   28325    &   19543         &     17589    &  1954      &      3.01         &   40             \\ \hline
All        & 17.7M      & 335.0k      &      248.8k      &    223.9k     &   24.9k     &    3.43           &  172               \\ \hline
\end{tabular}
\end{table}
The overall category-level data statistics are shown in Table~\ref{tab:dataset_statistics}. 
We provide the original raw data number, the number of data that have MP4 format videos, the filtered data statistics after our CLIP-MoE-based data curation, the numbers of training and evaluation data, average attribute counts, and the unique number of attributes for each category.
Overall, our proposed VideoAVE dataset covers 14 different domains with 172 unique attributes.
There are 223.9k training samples and 24.9k evaluation samples in total. 
The average attribute count for each product is 3.43.

\section{Experiment for Benchmarking}

\subsection{Experimental Setup}

\subsubsection{Data and Baselines Setup}
We randomly split the dataset into training and evaluation sets with a 9:1 ratio. 
Detailed dataset statistics are provided in Table~\ref{tab:dataset_statistics}.
To benchmark performance on VideoAVE, we evaluate four state-of-the-art video-language models: VideoLLaVA~\cite{lin-etal-2024-video}, VideoLLaMA3~\cite{damonlpsg2025videollama3}, InternVideo2.5~\cite{wang2025internvideo2}, and Qwen2.5-VL~\cite{yang2024qwen2}. Additionally, we fine-tune the attribute-conditioned best-performing model Qwen2.5-VL to assess its adaptability and performance gains on VideoAVE.

\subsubsection{Evaluation Setup}
We evaluate benchmarks on VideoAVE from two perspectives. 
The first focuses on attribute-conditioned value prediction, where the model is given a list of attributes and tasked with extracting their values from the product video. 
The second perspective is a more general setting of open attribute-value pair extraction, where no attributes are provided in advance. 
In general, the input for AVE is product video and the output is attribute-value pairs.
The prompt template is: \textit{Your task is to extract the attributes of the product in this video. Answer it in this format only: `attribute1': `attribute1\_value', `attribute2': `attribute2\_value'}.
Additionally, we present the results from both category and attribute levels.
Following prior works~\cite{zhang-etal-2023-pay, gong-etal-2025-visual}, we adopt fuzzy matching instead of exact match to accommodate natural language variations. 
A predicted value fuzzily matches a label when their common substring length exceeds half of the label length.
Category-level evaluation computes the Fuzzy F1 score across all attribute-value predictions for products within each category, with final scores averaged over all items.
Attribute-level evaluation assesses the Fuzzy F1 score separately for each attribute across the full dataset, offering insight into which attributes the model can handle more effectively.
All fine-tuning jobs are conducted on 8 A100 GPUs around 7 days and inference jobs require 1 A100 GPU. 

\subsection{Results}

\begin{table*}[]
\small
\centering
\caption{\label{tab:attribute_result} Attribute-level fuzzy F1 (\%) results. Bold indicates the best scores; gray highlights denote fine-tuned results.}
\begin{tabular}{lcccccccccccc}
\hline
\multirow{2}{*}{Model}                               & \multirow{2}{*}{Brand} & \multirow{2}{*}{Color} & \multirow{2}{*}{Material} & \multirow{2}{*}{\begin{tabular}[c]{@{}c@{}}Country \\ of Origin\end{tabular}} & \multirow{2}{*}{\begin{tabular}[c]{@{}c@{}}Item \\ Form\end{tabular}} & \multirow{2}{*}{\begin{tabular}[c]{@{}c@{}}Power \\ Source\end{tabular}} & \multirow{2}{*}{Shape} & \multirow{2}{*}{\begin{tabular}[c]{@{}c@{}}Suggested \\ Users\end{tabular}} & \multirow{2}{*}{Pattern} & \multirow{2}{*}{\begin{tabular}[c]{@{}c@{}}Finish \\ Type\end{tabular}} & \multirow{2}{*}{\begin{tabular}[c]{@{}c@{}}Mounting \\ Type\end{tabular}} & \multirow{2}{*}{\begin{tabular}[c]{@{}c@{}}Sport \\ Type\end{tabular}} \\
                                                     &               &                        &                           &                          &                            &                                                                          &                        &                        &                          &                                                                         &                                                                           &                                                               \\ \hline
 Video-LLaVA     &  6.49    &   19.62       &   15.12    &    15.61        &  13.02     &   2.31         &  22.89          & \textbf{2.28}      &  21.85     &  10.68       &    8.62        &   25.73      \\
 VideoLLaMA3     & 43.27  & 32.76    & 28.34 & 29.68      & 19.82 & 8.82      & 34.70      & 1.68 & 27.64 & 14.52   & 12.52      & 44.16 \\
 InternVideo2.5     &  42.41      &  \textbf{34.95 }       &   28.37    &   \textbf{43.78}         &  19.45    &  8.03         &    29.95       &  1.58    &   44.13    &   \textbf{17.47}      &    \textbf{22.21}        &  47.07      \\
 Qwen2.5-VL     &  \textbf{46.89}     &   34.42       & \textbf{33.69}      &  34.90          &  \textbf{25.34}    &  \textbf{11.96}         &    \textbf{35.71}       & 1.87      &  \textbf{46.16}   &  16.27     &    20.40       &   \textbf{49.58}      \\ 
  Qwen2.5-VL$^{*}$     &  \textbf{55.92}      &   \textbf{47.54}       & \textbf{58.15}     &  \textbf{65.73}          &  \textbf{63.38}     &   \textbf{75.31}         &    \textbf{68.19}        & \textbf{79.72}      &  \textbf{69.03}     &   \textbf{48.88}      &    \textbf{68.79}        &   \textbf{70.69}      \\
 \hline
\end{tabular}
\end{table*}

\subsubsection{Category-Level Results}
We present the domain-level fuzzy F1 scores across 14 categories for all evaluated models under both attribute-conditioned and generalized settings in Table~\ref{tab:main_result}.
In the attribute-conditioned setting, Qwen2.5-VL achieves the best performance among all zero-shot models.
In the more challenging generalized setting, where attribute prompts are not provided, overall performance drops, and InternVideo2.5 emerges as the strongest zero-shot model. 
We observe that: (1) Fine-tuning significantly improves performance, especially in the generalized setting. However, despite these gains, current state-of-the-art video-language models still have substantial room for improvement on the e-commerce AVE task. (2) The generalized setting is more difficult as models must not only predict accurate attribute values but also identify which attributes are relevant for a given product. This calls for more nuanced evaluation metrics besides F1 to better capture the model’s decision-making process in realistic e-commerce scenarios.

\subsubsection{Attribute-Level Results}
Table~\ref{tab:attribute_result} reports the attribute-level performance of all evaluated models. Due to space limitations, we present results for the top 12 most frequent attributes.
We observe that for several less common attributes (i.e., power source, suggested users, etc.), fine-tuning significantly improves attribute-conditioned performance.
For example, a pre-trained VLM may respond to suggested users with an overly descriptive or ambiguous answer when queried directly. However, the model better aligns predictions with a fixed candidate set (i.e., Mens, Womens, Unisex-Adults, Unisex-Teen, etc.) after fien-tuning. 
This indicates that for certain attributes, the pre-trained knowledge of VLMs is insufficient for accurate AVE, and finetuning is necessary to learn the mappings between attributes and their valid values.
Additionally, all models are still struggling with attributes that are inherently subjective (i.e., interpreting multicolor vs. blue and silver), not visually discernible (i.e., distinguishing material of plastic v.s. polyester), or highly sensitive to lighting (i.e., recognizing finish types of blushed v.s. polished).

\subsubsection{Modality-Level Results}
\begin{table}[]
\small
\centering
\caption{\label{tab:modality_result} Attribute-level performance (fuzzy F1) comparison among different modalities of text, image and video.}
\begin{tabular}{lcccc}
\hline
\multirow{2}{*}{Attributes} & \multirow{2}{*}{Title} & \multirow{2}{*}{\begin{tabular}[c]{@{}c@{}}Single\\ Image\end{tabular}} & \multirow{2}{*}{\begin{tabular}[c]{@{}c@{}}Multiple\\ Images\end{tabular}} & \multirow{2}{*}{Video} \\
                            &                        &                                                                         &                                                                           &                        \\ \hline
Caster Type                  & 9.52                   & 33.33                                                                   & \textbf{40.00}                                                                     & 38.46                  \\
               Fretboard Material Type                     &        10.14                &   31.35                                                                      &  42.17                                                                         &         \textbf{48.28}               \\
   Guitar Bridge System                                      &      1.50                  &      16.00                                                                   &  21.62                                                                         &      \textbf{29.30}                  \\
        Hair Type                    &    \textbf{31.69}                    &    1.79                                                                     & 1.82                                                                          &    25.41                    \\
        Pattern                            &      22.61                  &   48.96                                                                      &      \textbf{55.01}                                                                     &   46.16                    \\
        Rug Form Type                    &   65.62                     &   48.57                                                                      &  71.64                                                                         &      \textbf{83.54}                  \\
    Signal Format                        &      66.95                  &        66.41                                                                 &     71.90                                                                      &      \textbf{84.01}                  \\
  Strap Type                          &     2.74                   &    18.18                                                                     &  20.00                                                                         &   \textbf{22.47}                     \\
 Switch Type                           &    37.50                    &       35.29                                                                  &   32.26                                                                        &  \textbf{40.00}                      \\
 Towel Form Type                           &    \textbf{70.97}                    &       26.09                                                                  &  33.33                                                                        & 66.67                     \\
 \hline
\end{tabular}
\end{table}

\begin{figure}[htp] 
 \center{\includegraphics[height=4.4cm,width=8.2cm]{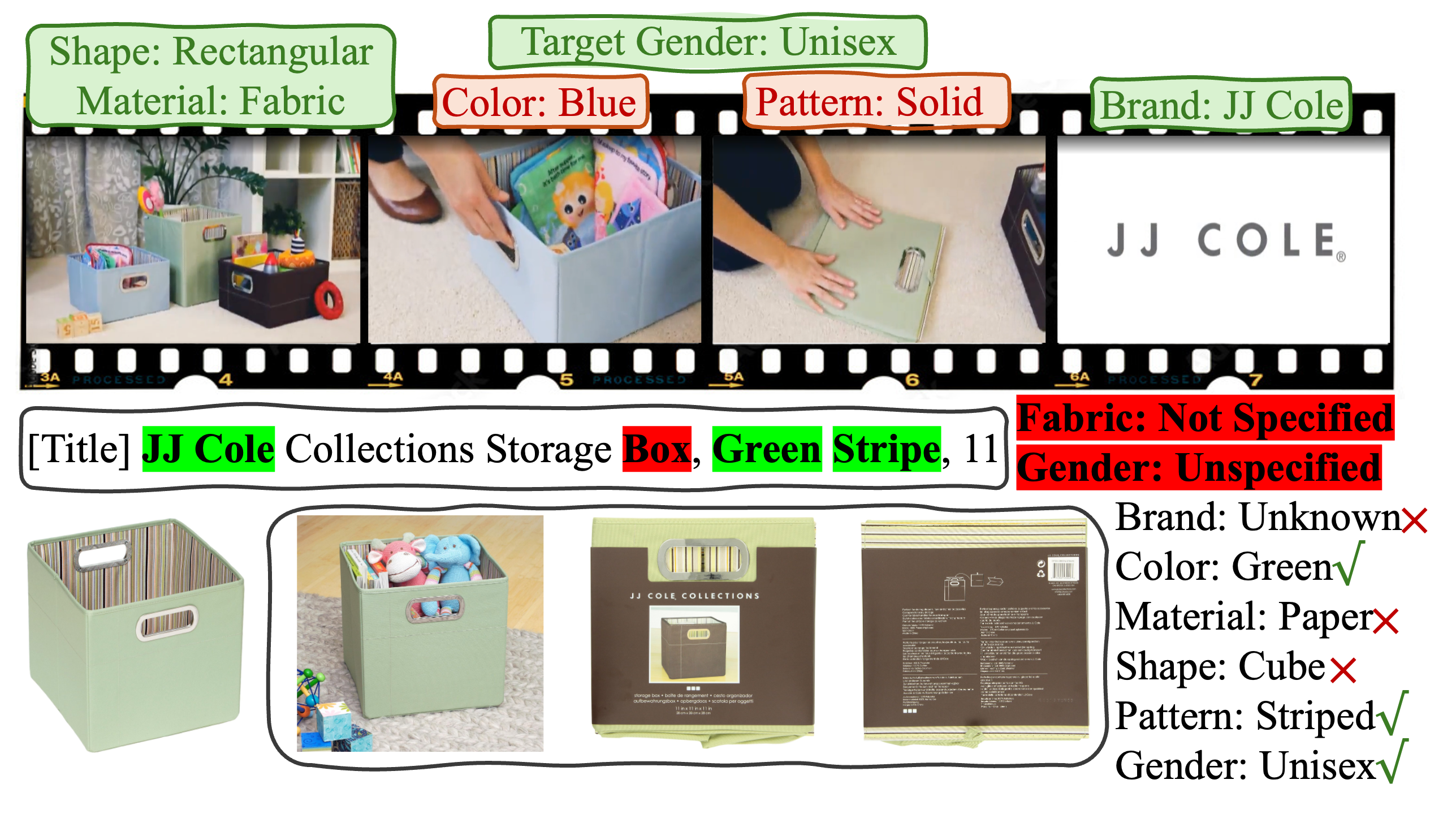}}
 \caption{\label{fig:error} Case study of AVE performance with text, multiple images, and video inputs.}
 \vspace{-4mm}
 \end{figure}

Table~\ref{tab:modality_result} presents a comparison of attribute-level fuzzy F1 performance across four input modalities: product title (text), single image, multiple images, and video.
Due to space constraints, results are shown for ten randomly selected attributes.
For a fair comparison, we use the same pre-trained VLM (Qwen2.5-VL) that supports all three modalities. All images used are high-resolution.
Overall, visual modalities consistently outperform text-only input, indicating the limitations of relying solely on textual descriptions for AVE. Among visual inputs, video achieves the highest performance for attributes that require fine-grained or spatial understanding (i.e., switch type, etc.), leveraging temporal and multi-angle cues not present in a static image.
However, some attributes perform better with textual input (i.e, hair type, towel form type), where values are visually ambiguous and better specified in the product listing.
Besides, some attributes (i.e., vaster type, pattern) benefit more from multiple images input than from video, which suggests that while videos offer rich visual content, they may also introduce irrelevant or noisy information that can hinder performance.
Figure~\ref{fig:error} shows an example where the rectangular shape and fabric surface of the storage boxes are accurately captured from the video. However, the presence of distracting visual elements (a blue box with a solid pattern) introduces noise into the video content, resulting in incorrect predictions for the color and pattern attributes.
These findings motivate future research toward more robust video understanding methods to extract the most relevant visual content within long product videos.

\section{Conclusion}
We introduce VideoAVE, the first video-to-text AVE dataset to address the limitations of existing datasets constrained to text-to-text or image-to-text settings.
To ensure high-quality data, we propose a post-hot CLIP-based Mixture of Experts filtering mechanism (CLIP-MoE) to remove inconsistent video-product pairs.
VideoAVE spans 172 unique attributes across 14 domains, comprising 224k training and 25k evaluation data.
We benchmark four state-of-the-art VLMs on VideoAVE under both attribute-conditioned and generalized AVE settings. We further analyze the impact of input modality (text, image, video) on performance. Our results highlight the challenges of robust video understanding in real-world e-commerce. 
For future work, we will explore methods for identifying the most relevant visual content to enhance both robustness and inference efficiency.

\bibliographystyle{ACM-Reference-Format}
\bibliography{sample-base}


\end{document}